\documentclass[journal ]{new-aiaa}

\usepackage[utf8]{inputenc}
\usepackage{textcomp}
\usepackage{soul}
\usepackage{graphicx}
\usepackage{amsmath}
\usepackage[version=4]{mhchem}
\usepackage{siunitx}
\usepackage{longtable,tabularx}
\usepackage{color}
\usepackage{xcolor}
\usepackage{soul}
\usepackage{rotating}

\title{Simulating Integration of Urban Air Mobility into Existing Transportation Systems: A Survey}

\author{Xuan Jiang \footnote{Ph.D. Candidate, Department of Civil and Environmental Engineering, University Avenue and, Oxford St, Berkeley, CA 94720} and Yuhan Tang\footnote{Master Student, Department of Civil and Environmental Engineering, University Avenue and, Oxford St, Berkeley, CA 94720} and Junzhe Cao\footnote{AIAA Member, Undergraduate Student Researcher, Department of Civil and Environmental Engineering,University Avenue and, Oxford St, Berkeley, CA 94720, j.cao@berkeley.edu}}
\affil{University of California, Berkeley, Berkeley, CA 94720}

\author{Vishwanath Bulusu, Ph.D \footnote{Aerospace Research Scientist, Crown Consulting, Inc., Moffett Field, CA 94035}}
\affil{Crown Consulting, Inc., Moffett Field, CA 94035}

\author{Hao (Frank) Yang, Ph.D \footnote{Research Assistant Professor, Department of Civil and Environmental Engineering, Latrobe Hall, 3400 N Charles St \#205, Baltimore, MD 21218}}
\affil{Johns Hopkins University, Baltimore, MD 21218
}

\author{Xin Peng, Ph.D.\footnote{Researcher, Department of Civil and Environmental Engineering, University Avenue and, Oxford St, Berkeley, CA 94720}}
\affil{University of California, Berkeley, Berkeley, CA 94709}

\author{Yunhan Zheng, Ph.D \footnote{Researcher, Department of Civil and Environmental Engineering, 77 Massachusetts Avenue Cambridge, MA 02139} and Jinhua Zhao, Ph.D \footnote{Professor, Department of Urban Studies and Planning, 77 Massachusetts Avenue Cambridge, MA 02139}}
\affil{Massachusetts Institute of Technology, Cambridge, MA 02139}

\author{Raja Sengupta, Ph.D \footnote{Professor, Department of Civil and Environmental Engineering, University Avenue and, Oxford St, Berkeley, CA 94720}}
\affil{University of California, Berkeley, Berkeley, CA 94720}

\begin{document}

\maketitle

\section{Introduction}


\lettrine{U}{rbanization} projects indicate that by 2050, more than 70\% Europeans and 80\% North Americans will call cities home. In developed regions, where nearly 80\% of the population is expected to be urbanized \cite{leeson2018growth}, such rapid growth has also brought a host of problems including issues with mobility, infrastructure, congestion, and pollution\cite{chen2008sustainable}. In addition to harming people and the environment, these problems also incur substantial financial cost \cite{litman2009transportation}. In Europe alone, congestion costs an estimated 130 billion euros per year \cite{newbery2016benefits}. Major cities such as New York, London, Paris, and Tokyo have already reported significant economic losses due to excessive fuel and vehicle operating costs \cite{kenworthy1999patterns, byrne2019multivariate}. Innovative approaches to urban transportation are therefore necessary to address these issues.

Urban Air Mobility (UAM) is one such concept that could potentially revolutionize the way we move within different areas. UAM refers to the use of air vehicles for transporting people and cargo in urban environments \cite{mazur2022regulatory}. By leveraging technological advancements in batteries, electric propulsion, and vertical take-off and landing (eVTOL) capabilities, UAM could enable point-to-point flights and bypass ground congestion \cite{goyal2018urban}. Companies like Joby, Wisk, Electra and  Lilium are at the forefront of developing novel UAM aircraft, while Zephyr Airworks, Airbus, and Volocopter GmbH have already conducted extensive test flights with their eVTOL demonstrators in various countries around the world, including the United States, Japan, Singapore, New Zealand, France, and India \cite{rajendran2019insights}. Successful efforts could lead to the mass production of such vehicles and the development of air taxi services, disrupting not just aviation but also mobility systems and urban planning.

 The innovative business models of UAM could make it a candidate alternative for providing high-speed, user-oriented mobility services that complement and enhance existing transportation options. For example, UAM has gained significant attention in recent years as a potential solution for addressing mobility challenges such as congestion in metropolitan areas \cite{pons2022understanding, straubinger2020overview} and improving transportation efficiency and accessibility \cite{viswanathan2022challenges}. However, its integration into existing urban transportation systems is still in the early stages of development \cite{rothfeld2021potential}. UAM integration is a complex task that requires a thorough understanding of its impact on traffic flow and capacity \cite{niklass2020collaborative}. There are many factors to be addressed including infrastructure requirements, regulation, safety, and public acceptance \cite{cohen2021urban}. Traffic simulation is a useful tool for understanding such impact of UAM in metropolitan areas. Simulation allows researchers to test different scenarios and analyze the consequences of UAM integration in a virtual controlled environment \cite{rothfeld2018agent, rothfeld2021potential}. In this paper, we survey the current state of research on UAM integration in metropolitan-scale traffic using simulation techniques. We also discuss how simulation can assist with evaluating the potential benefits of UAM, such as reduced travel times and improved accessibility for underserved areas. 
 This research endeavors to provide a comprehensive and nuanced analysis of simulation capabilities needed and available tools to be leveraged for studying the key considerations surrounding the integration of UAM in urban transportation. 

Unlike some existing overview papers addressing the UAM concept \cite{johnson2022nasa, straubinger2020identifying, rajendran2020air, cohen2021urban, garrow2021urban}, we seek to complement the knowledge base by undertaking a meticulous examination and synthesis of the most consequential sources of UAM development and integration research, which employs a simulation approach for greater depth of understanding. We commit to a detailed appraisal of the present state-of-the-art and future trajectory of various aspects of UAM simulation, leveraging insights from both academic literature and pragmatic examples. Through this methodological approach, we intend to inform the selection of appropriate simulators for UAM-centric research, thereby facilitating future scholarly endeavors in this dynamic field. Our findings are organized in two parts: First, we delineate a comprehensive array of simulators related to UAM integration research. This will provide a foundation for understanding the breadth of tools available, their specific functionalities, and their application contexts. Second, we discuss how these simulators can be effectively deployed to serve diverse research topics within the UAM landscape. This will equip researchers with pragmatic strategies for leveraging these tools in their own explorations, thereby fostering further advancements in the field. Our ultimate goal is to provide a resource that is both comprehensive in its presentation of available simulators and practical in its guidance for their application in UAM research.

This paper is organized as follows: The "Methodology for Systematic Literature Review" section delineates our rigorous approach to literature search and screening. In "Significant Discoveries from the Comprehensive Search," we spotlight key findings from the literature, encompassing advancements in simulation and emerging UAM integration research. "Key Research Gaps and Future Directions" highlights the areas we identified as underserved in current studies and proposes potential avenues for exploration. We conclude in "Challenges and Conclusions," reflecting on the hurdles encountered and the insights derived from our comprehensive analysis.

\section{Methodology for Systematic Literature Review}

The main goal of this search was to find the studies on what is the latest update on UAM development and integration research, through employing a simulation approach for greater depth of understanding, and what are the features of each simulator that is used to simulate UAM. The following search terms were applied:
{\small
\begin{align*}
&\left(
    \begin{array}{l}
        \text{"UAM" OR "Air Taxi" OR} \\
        \text{"Air-Taxi" OR "Urban Air Mobility" Or "AAM"} \\
        \text{AND} \\
        \text{"Simulation" OR "Simulate" OR} \\
        \text{"Simulator"} \\
        \text{AND} \\
        \text{"Integrate" OR "Integration" OR "Multi"} \\
    \end{array}
\right) \\
&\text{OR} \\
&\left(
    \begin{array}{l}
        \text{"Simulation" OR "Simulate" OR} \\
        \text{"Simulator"} \\
        \text{AND} \\
        \text{"Transportation" AND "Transit"} \\
    \end{array}
\right)
\end{align*}
}

In our research on UAM, often termed "Air Taxi," we initially outlined the field's broad research scope and then conducted specific searches on simulators. Our methodology involved filtering results to focus on papers discussing simulation in titles, abstracts, or keywords, and then narrowing down to topics of "integration" into transportation networks and "multi-modal" approaches involving UAM. We used Scopus as our data source, emphasizing English-language publications as of September 9, 2023.

Our comprehensive search yielded 2,638 articles, which were further shortlisted. We excluded papers focusing solely on terrestrial transportation and those dealing with Geographic Information Systems (GIS), Global Positioning System (GPS) in UAM conflict resolution, and economic aspects of UAM but didn't discuss things related to simulations. Realizing the need to cast a wider net to capture the full lists of UAM literature, we explored additional platforms such as Google Scholar and the journals of the American Institute of Aeronautics and Astronautics (AIAA). Engaging with subject matter experts, our objective is to improve our understanding and ensure that no significant study slips through our analytics. Some papers were also manually added due to limited representation in our initial data set. 

After screening, 60 papers were selected for detailed analysis. These were categorized into two main domains: technical aspects of UAM simulation and four thematic clusters - impact on existing traffic and on-demand services, safety and risk assessment, equity influence, and economic and environmental impacts. This categorization provides a foundation for further discussion and analysis in the UAM literature.

\section{Significant Discoveries from the Comprehensive Search}

\subsection{Assessing the Implications of UAM Integration}
\begin{figure}[!htbp]
\centering
\centerline{\includegraphics[width=1\linewidth]{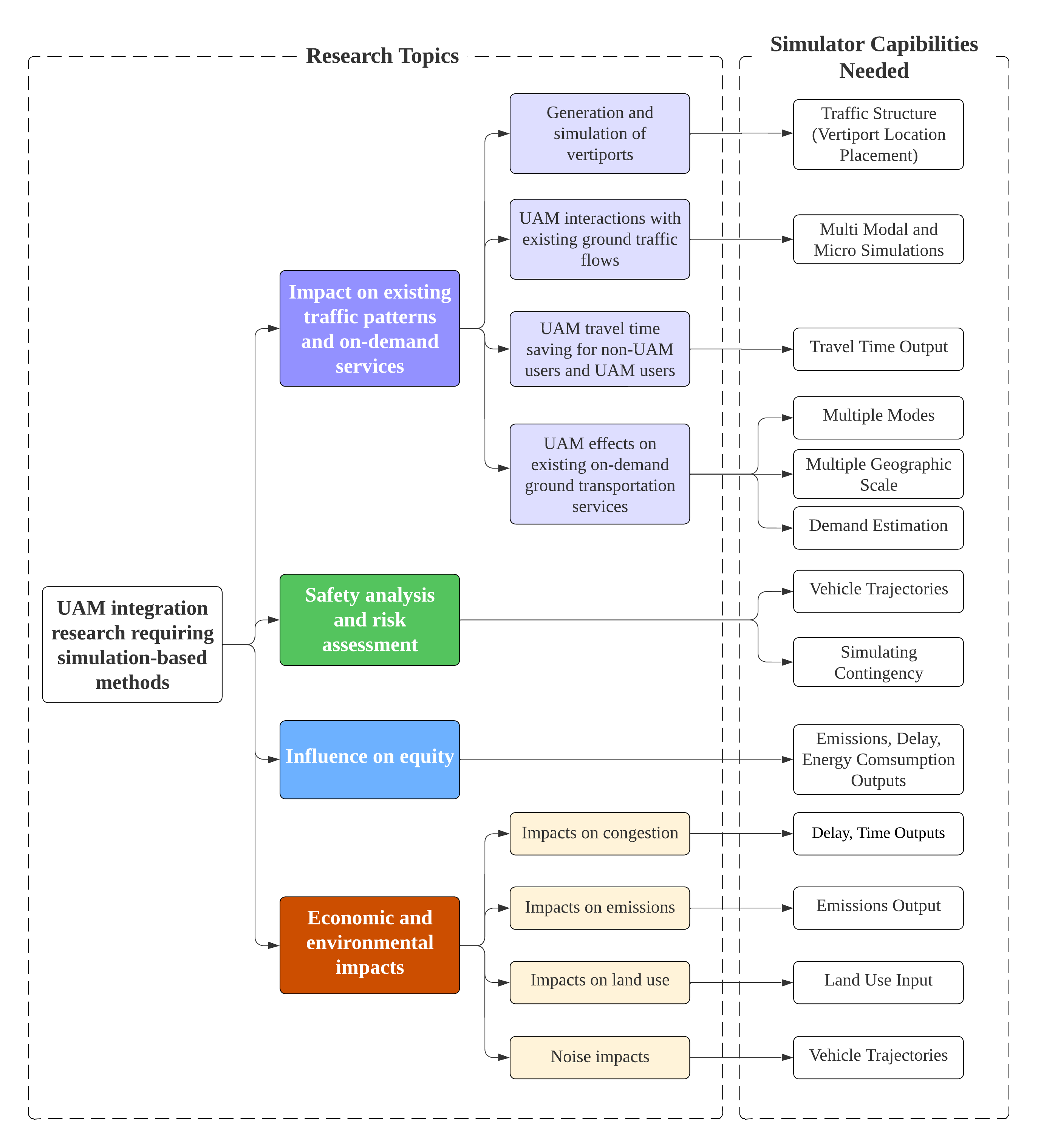}}
\caption{UAM Integration Study with Simulation Research Fields}
\label{UAM integration study with simulation research fields}
\end{figure}

There is a critical need to explore UAM integration into existing transportation systems. UAM could redefine mobility in urban landscapes by leveraging technological advancements to address changing mobility patterns, transportation needs, and environmental concerns. Simulation, as a research tool, holds immense value in UAM studies. It allows one to model complex systems, predict outcomes, and facilitate decision-making. Researchers worldwide are harnessing the power of simulation tools to explore areas such as safety, efficiency, and environmental impacts. We, therefore, conducted a comprehensive review  of research on UAM integration challenges and relevant simulation-based tools and methods. This included scholarly articles, project reports, policy documents, and information on prototypes, demonstrations, and current experiments in the field of UAM. We accessed a variety of academic databases including IEEE Xplore, Scopus, EBSCO, ScienceDirect, Web of Science, and Google Scholar. From the review, we identified key areas for UAM integration research, established problems to be studied and derived the simulation capabilities needed as shown in Figure \ref{UAM integration study with simulation research fields}. 

The framework elucidates an assortment of research topics relevant to the integration of UAM through the employment of simulation studies. Each topic is accompanied by its respective simulator requirements. The topics span from multi-modal capability and micro-simulation, which delve into the labyrinth of interconnected transportation networks, to the examination of alterations in prevailing traffic patterns of consumers and other modes caused by the introduction of UAM. They also encompass the safety analysis and risk assessment, which scrutinizes the potential hazards and safety implications of UAM deployment, as well as its influence on equity, examining whether UAM's introduction could inadvertently exacerbate or mitigate social inequalities. Furthermore, the framework includes studying the dynamics of UAM vehicle interactions with other vehicular systems and evaluating the gains or setbacks in travel time associated with UAM. The simulator capabilities needed are diverse, ranging from the provision for detailed travel time output and the capability to simulate contingency scenarios, to features that model vehicle trajectories. We also include considerations for reduced emissions, enhanced land use, and noise impacts. In the subsequent sections, we delve into these themes in detail, providing a comprehensive survey that we hope will serve as a valuable resource for researchers and practitioners in the field.

\textbf{Optimized Vertiport Generation and Traffic Analysis:} Simulation methods play a pivotal role in determining the optimal distribution of vertiports, which are crucial for UAM service efficiency. These simulations encompass clustering algorithms, GIS-based analysis, and integer programming models. For instance, in the study by Lim \& Hwang \cite{lim2019selection}, the k-means algorithm was used to determine vertiport locations along busy routes, effectively reducing congestion and promoting UAM adoption. \citet{arellano2020data} employed GIS-based analysis, considering factors like accessibility and cost-effectiveness to identify potential vertiport locations. Their data-driven approach was showcased in a Munich case study. Furthermore, \citet{wu2021integrated} used integer programming to minimize travel expenses for all users, optimizing vertiport locations and reducing congestion, demonstrating the flexibility of simulation in addressing UAM network design challenges.

\textbf{Safety Assessment and Risk Mitigation: }Simulation is instrumental in ensuring the safety of UAM operations through comprehensive assessments. Various models and methods, such as collision risk modeling and hazard assessments, are employed. \citet{borener2015modeling} utilized Monte Carlo simulations and negative binomial regression to analyze the safety of UAM operations under different scenarios, considering communication failures and weather conditions. Additionally, \citet{bijjahalli2022unified} proposed a risk management framework that factored in severe weather conditions and collision risk modeling to assess UAM safety. These studies underscore the critical role of simulation in identifying potential hazards and enhancing the safety of UAM systems.
Equity Considerations: Simulation techniques are instrumental in addressing equity concerns associated with UAM integration. \citet{chen2022analysis} conducted agent-based simulations to assess potential equity gaps in UAM adoption among different income groups. His findings shed light on how UAM may initially benefit high-income users but ultimately contribute to reduced equity gaps in the long term. \citet{tang2021automated} utilized integer programming to promote fairness among service companies, ensuring that UAM benefits are distributed equitably. The ability of simulation to model and analyze the impact of UAM on various socioeconomic groups is crucial for equitable urban transportation planning and policy decisions.

\textbf{Economic Benefits:} Simulation enables the assessment of economic advantages, such as reduced travel costs and increased efficiency. Peng \textit{et al.}\cite{peng2022hierarchical}proposed an on-demand multi-modal UAM service that considered passenger needs and allocation, demonstrating how simulation can optimize UAM services to adapt to changing demand. This approach enhances the overall efficiency of the UAM system, reducing travel costs for users. Furthermore, \citet{yedavalli2019assessment}conducted surveys to understand potential users' concerns about equity and economic access to UAM services. Their findings inform strategies to ensure UAM remains an economically viable option for a wide range of users.
Environmental Improvements: Simulation plays a crucial role in assessing environmental benefits, particularly the reduction of greenhouse gas emissions and air pollution. \citet{Yedavalli2021b}used Microsimulation Analysis for Network Traffic Assignment (MANTA) simulator to analyze how vertiport placement could affect traffic patterns. By expanding vertiport accessibility, the efficiency of UAM services can be improved, potentially reducing the number of vehicles on the road and, consequently, emissions. Additionally, \citet{wu2021integrated} investigated the impact of UAM services on traffic congestion. \citet{mudumba2021modeling} from Purdue University show that the eVTOL aircraft are more environmental friendly in their research. Their findings suggested that UAM services could decrease the number of vehicles on the road, resulting in reduced congestion and potentially lower emissions.

\subsection{Conceptual Framework for Evaluating Simulators in UAM Integration}
The integration of UAM into existing transportation systems holds the potential to redefine urban mobility and significantly impact traffic patterns. However, understanding these impacts requires comprehensive research, particularly in the realm of simulation. The ability to accurately simulate the introduction of UAM into current traffic systems is paramount to foreseeing potential issues, planning suitable infrastructure, and ensuring efficient integration. As cities continue to grapple with increasing traffic congestion and the associated economic and environmental repercussions, the investigation into UAM's impact on existing traffic patterns becomes even more crucial. We mainly focus on the following four parts.

First, The integration of UAM into existing transportation systems could significantly impact traffic patterns. The ability to accurately simulate the introduction of UAM into current traffic systems is paramount to foreseeing such potential issues, planning suitable infrastructure, and ensuring efficient integration. 

Second, the prospect of UAM inevitably brings potential risks and safety concerns associated with its operation. From technical malfunctions to accidents, UAM presents unique safety challenges that need to be assessed and mitigated. Comprehensive simulation studies are vital in anticipating these risks and developing effective countermeasures, thereby ensuring the safety of UAM operations and the broader public.

Third, UAM also holds the potential to enhance accessibility and reduce travel times, particularly for underserved communities. However, it is essential to investigate whether the benefits of UAM will be distributed equitably among all segments of the population. This research into UAM's influence on equity can provide valuable insights to policymakers and stakeholders, ensuring that the deployment of UAM promotes inclusive urban mobility.

Fourth, UAM could herald significant economic and environmental benefits. From reducing congestion-related costs to potentially lowering emissions, UAM's economic and environmental advantages are tantalizing. However, these benefits need to be quantitatively assessed and balanced against potential costs. Simulation studies can provide critical insights into the long-term economic viability and environmental sustainability of UAM, guiding strategic decisions and policies.

\begin{sidewaysfigure}
  \centering
  \includegraphics[width=\textwidth]{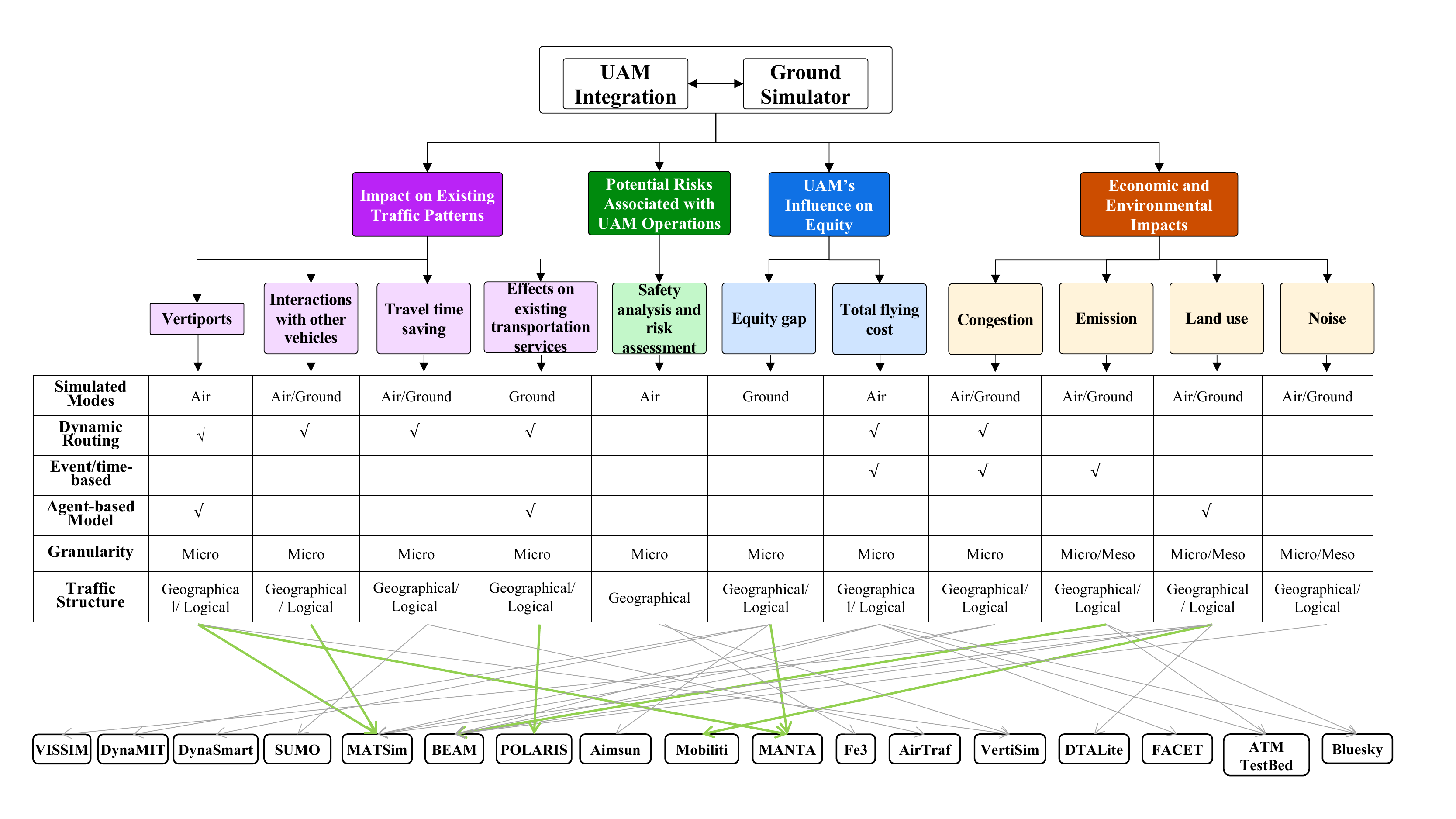}
  \caption{Research Usage of the Survey at a Glance}
  \label{Resarch Usage of the Paper at a glance}
\end{sidewaysfigure}

Considering the four parts, Figure \ref{Resarch Usage of the Paper at a glance} captures the comprehensive utilization guidance for our study. It serves as a navigation tool, enabling users to effortlessly identify the research requirements corresponding to their specific research topics of interest. 
 In addition, this figure provides visibility into the current simulation landscape by highlighting, with a distinctive green line, the simulation tools that researchers are presently employing in the field. Furthermore, to assist in the informed selection of simulation tools, the figure directs users' attention to recommended simulators, marked by prominent grey arrows. These recommendations are borne out of our extensive analysis and understanding of the capabilities and advantages of various simulation tools in the context of UAM research. 
 For instance, there are existing researchers using Mobiliti to figure out economic and environmental impacts such as land use \cite{fujimoto2017parallel}, which is shown as the green line in Figure \ref{Resarch Usage of the Paper at a glance}. However, after analyzing the simulated modes, routing, model type, granularity, and traffic structure of land use analysis, VISSIM, MATSim, BEAM, and DTALite also have the ability to do such things, which is shown in the grey line.
 Thus, the figure not only maps the current state of affairs but also offers guidance for future research endeavors, fostering a robust, informed, and effective approach to UAM studies. 

\subsection{Variations in the Capabilities of Current Simulators}
The diverse array of current simulators has been comprehensively compared across 10 ground simulators and 7 air simulators, as illustrated in Table \ref{A Comparison Between Different Ground Simulators (Part 1)}, \ref{A Comparison Between Different Ground Simulators (Part 2)}, \ref{A Comparison Between Different Air Simulators (Part 1)} and Table \ref{A Comparison Between Different Air Simulators (Part 2)}. This section will sequentially examine why we chose them, their unique functions, and their relationships.

\subsubsection{Multi-modal and Dynamic Routing - Two Key functionalities}
Multi-modal simulators are widely used in urban simulations. In urban  VISSIM utilizes the car-following model \cite{fellendorf2010microscopic}, and caters to various urban and highway scenarios. VISSIM handles multimodal simulation, including pedestrian modeling based on the Social Force Model \cite{helbing1995social}, influenced by a mix of social, psychological, and physical elements. Its key strength is traffic demand modeling via OD patterns and dynamic vehicle assignment, useful for combined UAM and ground vehicle simulation \cite{williams2007simulation}. 

While VISSIM can perform multi-modal simulation unlike DynaMIT, the latter has pioneered dynamic routing in simulations \cite{ben2002real}. DynaMIT, developed by \citep{ben2002real}, is designed to intervene in the behavior of individual vehicles as they move through a transportation network, and it can be used to study a wide range of traffic management and control strategies. It includes tools for modeling the interactions between vehicles, pedestrians, and other traffic, as well as tools for analyzing traffic flow and predicting the impacts of different management strategies. However, DynaMIT has limited model geometry modifications and lacks structured technical support, relying on online user groups. Without dynamic routing, VISSIM struggles with complex urban systems with multimodal traffic demands. DynaSmart pioneers dynamic rerouting in multimodal tranportation simulators, modeling static network elements and dynamic vehicle behaviors \cite{zhao2017dynamic}. Its strengths include accurate real-time traffic patterns and network performance modeling, enabled by dynamic traffic assignment and traffic control strategies \cite{mahmassani1998dynamic}. 

SUMO and Aimsun combine the advantages of both multi-modal and dynamic routing. With adequate resources, a desktop PC can simulate up to 200,000 vehicles in real time, encompassing various traffic aspects.  However, their disadvantages involve potential performance issues in large-scale simulations, complexity for new users, and variable quality of documentation and support as an open-source project. Aimsun allows users to model and analyze vehicular and pedestrian movements within transportation networks \cite{casas2010traffic}. Its comparison with SUMO, by \cite{ronaldo2012comparison}, showed high reliability in replicating traffic dynamics but lower computational efficiency. Aimsun uses microsimulation, modeling individual vehicle and pedestrian behaviors in detail, compared to VISSIM's approach based on road geometry, traffic demand, vehicle characteristics, driver behavior, and signal timing.

\subsubsection{Open Source}
The significance of open-source software in traffic simulation, such as SUMO, lies in its transparency and customization, circumventing the limitations of proprietary solutions \cite{behrisch2011sumo}. SUMO, a pioneer in open-source, space-continuous simulation, facilitates individual vehicle routing and interaction parameters setting. 

After SUMO, many simulators are open-sourced, like MatSim, BEAM, POLARIS, and MANTA, etc. Open-sourced simulators increase the possibility for researchers to develop new simulators based on the old ones. Also, they stimulate the compatibility between different simulators. The iteration and development of simulators also boom due to open source. 

\subsubsection{Agent-based simulators}
As the scenario in simulation becomes more and more complex, agent-based models are invented to understand the effects of emerging technologies and urban development, by considering human behaviors and preferences. Agent-based simulators are computational models that use agents —representing vehicles, drivers, pedestrians, or infrastructure elements —to simulate and analyze their actions and interactions within the system. These agents are programmed with specific characteristics and decision-making processes, allowing them to respond to their environment and each other. This bottom-up modeling approach captures the complex dynamics of transportation systems, revealing emergent patterns and outcomes essential for analysis and planning. As a notable pioneer in an agent-based model, MATSim performs an activity-centric, adaptable Java-based framework, allowing for parallel processing, efficient management of large scenarios, and individual demand optimization \cite{w2016multi}.

BEAM, an expansion of MATSim, is a mesoscopic agent-based travel demand simulation framework that includes multimodal simulations and dynamic routing capabilities\cite{sheppard2017modeling}. It introduces a power threshold span for electric vehicles and incorporates innovative features such as a ride-hail manager and freights, offering greater authenticity and depth in simulations\cite{sheppard2016cost}. 

POLARIS is another example of agent-based modeling software. It integrates models usually managed by separate software, such as dynamic traffic assignment and disaggregate demand models, into a single process, overcoming longstanding integration issues and enhancing efficiency. Furthermore, POLARIS uses a custom memory allocator and parallelized discrete event engine, bolstering performance, and C++ binding database technologies for improved data input/output efficiency. However, the complexity of integrating multiple models, the technical skill required to use its specialized memory allocators and parallelized engines, and reliance on the quality and granularity of input data are POLARIS's drawbacks compared to other agent-based simulators.

Empowered by a supercomputer, Mobiliti is an agent-based and scalable transportation system simulator. It utilizes parallel discrete event simulation on high-performance computing platforms. By instantiating a vast number of nodes, links, and agents, Mobiliti successfully emulates the mobility patterns of a population navigating the road network in the San Francisco Bay Area. This simulation framework facilitates the assessment of congestion levels, energy consumption, and productivity depletion associated with the transportation system \cite{chan2018mobiliti}.

\subsubsection{Parallelism}
In order to get similar computing speed, supercomputer isn't a must. MANTA also performs at great speed not by the CPU, but by GPU parallelism.
With the objective of resolving the balance between augmenting the intricacy of a model and diminishing computational efficiency, a meticulously parallelized GPU implementation has been crafted to execute large-scale simulations in metropolitan areas swiftly, completing them within a matter of minutes. The computational efficiency achieved greatly enhances the forefront of large-scale traffic microsimulation. One characteristic that lends MANTA to GPU utilization is its discrete-event simulation approach which handles each vehicle and each timestep of simulation as separate but potentially concurrent events. This approach is highly parallelizable, making it ideal for GPUs, which are designed to handle thousands of threads simultaneously. Also, MANTA employs a unique "traffic atlas" for spatial distribution of vehicles, which maps vehicles and road segments directly into the memory, enabling rapid access and updates. This structure is highly amenable to the parallel processing capabilities of GPUs, allowing quick computations of vehicle positions and speeds over large networks. The runtime performance of MANTA is optimized through the integration of a distributed CPU-parallelized routing algorithm and a massively parallelized GPU simulation. This innovative approach leverages a unique traffic atlas, enabling the efficient mapping of vehicle distribution in memory as contiguous bytes \cite{yedavalli2021microsimulation}.

\subsubsection{Air simulators}
For the simulators that are only developed for air simulation, they also follow the categories that we've mentioned above. The Flexible engine for Fast-time evaluation of Flight environments (Fe$^{3}$) is a tool that allows stakeholders to analyze and study the impacts of high-density, high-fidelity, low-altitude traffic systems without the need for costly and logistically challenging flight tests \cite{xue2018Fe3}. It's the pioneer in GPU-parallel in air simulators. However, it may still demand significant computational resources, especially when applied to extraordinarily complex or large-scale scenarios. This aspect could pose challenges in terms of resource allocation and computational feasibility.

AirTraf, an air traffic simulator, operates as a component within the broader framework of the European Center HAMburg general circulation model (ECHAM) and Modular Earth Submodel System Atmospheric Chemistry (EMAC) model \cite{yamashita2020newly}. This integration allows for a comprehensive analysis of aviation's impact on climate through advanced simulation of aircraft routes and their associated emissions. AirTraf specifically aims to optimize aircraft routing to minimize environmental impact while taking into account operational factors such as economic costs and flight duration. In its more advanced form, AirTraf 2.0, the model significantly expands its capabilities by introducing multiple new routing options that target different optimization objectives such as minimizing contrail formation, reducing climate impact, and considering economic costs. This version of AirTraf allows for a detailed analysis of the trade-offs between these objectives, enabling the evaluation of routes that might, for example, reduce climate impact at a slight increase in operating costs. AirTraf operates by simulating every flight trajectory for the chosen routing options, factoring in daily changing atmospheric conditions, thereby providing a dynamic tool for understanding and potentially mitigating the aviation sector's impact on climate change. 

Vertiports are facilities designed to handle the landing, taxiing, parking, loading, and unloading, charging, repairing, and takeoff of aircraft, as well as the movement of passengers and cargo. These facilities are designed to provide all necessary infrastructure within a limited land area, which can limit their throughput capacity. As a result, vertiports may act as bottlenecks within the UAM system, limiting the overall capacity of the system. It is important for vertiports to be designed and managed efficiently in order to maximize their capacity and support the smooth operation of the UAM system \cite{vascik2019development}. In order to simulate vertiports, VertiSim has been developed. It is a tool that simulates the operations of a vertiport. It consists of three major components: the vertiport layout designer, the flight generator, and the vertiport manager \cite{yedavalli2021simuam}.

DTALite is an air traffic simulation software developed by the Federal Aviation Administration (FAA) to model and analyze the National Airspace System (NAS). It is used to predict the impact of changes to the NAS on capacity, efficiency, and safety, and to evaluate the performance of new technologies and procedures \cite{zhou2014dtalite}. DTALite uses real-world data on air traffic patterns, aircraft performance, and weather conditions to create simulations of the NAS. It can model different scenarios, such as changes to airspace management procedures or the introduction of new technologies, to predict the impact on capacity, efficiency, and safety. DTALite can also be used to evaluate the performance of new technologies and procedures, such as NextGen, the FAA's modernization program for the NAS.

FACET can model system-wide en route airspace operations over the contiguous United States, including aircraft trajectories, using round-earth kinematic equations. The software also enables the simulation of aircraft climb, cruise, and descent phases according to their individual aircraft-type performance models. Data such as climb/descent rates and speeds, cruise speeds, heading, airspeed, and altitude-rate dynamics are obtained from table lookups \cite{bilimoria2001facet}.

Utilizing the ATM TestBed, NASA aspires to expedite the NAS transformation by facilitating intricate simulation and evaluation of several integrated technologies, thereby testing NAS-wide operational solutions. This methodology enables a convincing demonstration of the potential benefits and feasibility of these concepts \cite{robinson2017development}. Consequently, the ATM TestBed has been used for early UAM concept demonstration simulations in the Dallas Fort Worth metroplex area. 

Technically, BlueSky is extendable via self-contained plugins, providing a means for users to customize and augment its functionality to suit their specific needs. BlueSky is inclusive of open-source data on navaids, performance data of aircraft, and geography, as well as global coverage navaid and airport data and simulations of aircraft. This extensive array of features and data positions BlueSky as a versatile asset for a broad range of air traffic research and analysis tasks \cite{hoekstra2016bluesky}.

The Aviation Environmental Design Tool (AEDT) is developed by a consortium of experts to evaluate and predict the environmental impacts of aviation activities \cite{roof2007aviation}. It integrates existing noise and emissions models to provide a comprehensive assessment of the interdependencies between aviation-related noise, exhaust emissions, performance, and economic parameters. AEDT enables stakeholders to make informed decisions regarding aviation policies and practices to minimize environmental impacts with future aircraft designs and technological scenarios.

The Aviation Emissions Inventory Code (AEIC) is a tool designed to rapidly estimate global emissions from scheduled civil aviation, incorporating the complexity of flight operations and modeling with a focus on quantifying uncertainty \cite{simone2013rapid}. AEIC simplifies the computational intensity required by other models, thus enabling quicker simulations that still accommodate various scenarios and uncertainty assessments through Monte Carlo simulations. Also, AEIC notably addresses emissions during all phases of flight, from takeoff to landing, and is capable of processing vast amounts of data to generate annual global emissions figures. It is instrumental for environmental impact assessments, policy analysis, and academic research, offering a valuable resource for understanding and managing the environmental impacts of aviation.

Some researchers come up with self-developed simulators when they are dealing with transportation modeling problems. In the paper about demand and capacity modeling for UAM by \cite{alvarez2021demand}, they developed a discrete-event simulation model to analyze the performance of AAM traffic networks. This model simulates the operations within a network of vertiports and aircraft by treating each aircraft and vertiport as individual resources with fixed capacities—seats in aircraft and vertipads at vertiports, respectively. The simulation progresses through discrete events, moving from one significant event to the next and assumes no change in the system state between events. It queues excess requests for these resources when demand exceeds capacity, handling them on a first-come, first-served basis. This approach enables the simulation to realistically model the flow of aircraft and passenger traffic, incorporating factors such as service times for loading and unloading passengers and holding times at vertiports. By configuring variables like fleet size, aircraft speed, vertiport locations, and the number of vertipads, the simulator allows users to test different scenarios, assess the impact of various network configurations on operational metrics such as delays and utilization, and optimize the UAM network's design to meet specific performance targets.

\subsection{Air traffic's interaction with ground traffic}
We shouldn't separate the results of ground and air simulators, especially in the situation of integrating UAM into existing ground transportation systems. In the UAM extension for MATSim, Maciejewski \textit{et al.} \cite{maciejewski2016dynamic} first declared the UAM Infrastructure, Vehicle, and Network Modeling. In the current implementation of the UAM network, UAM vehicles are able to navigate through airspace by following predefined air routes and waypoints, similar to the way that vehicles on the ground follow roads and intersections. These air routes and waypoints are referred to as links and nodes, respectively, and are organized into a network similar to a road network. Each UAM station is represented by two nodes: a ground-access node and a flight-access node, which are connected by a short link called the station link. The station link serves as the anchor for the UAM station and should be designed to minimize travel time. This implementation allows UAM vehicles to move through the network in a similar way to how ground vehicles move through a road network.

\begin{table}[]
\caption{A Comparison Between Different Ground Simulators (Part I)}
\label{A Comparison Between Different Ground Simulators (Part 1)}
\centering
\scalebox{0.8}{
\begin{tabular}{cccccc}
\hline
\textbf{Simulator} &
  \textbf{1. Vissim \cite{fellendorf2010microscopic}} &
  \textbf{2. DynaMIT \cite{ben2002real}} &
  \textbf{3. DynaSmart \cite{mahmassani1992dynamic}} &
  \textbf{4. SUMO \cite{behrisch2011sumo}} &
  \textbf{5. MATSim \cite{w2016multi}} \\ \hline
\textbf{\begin{tabular}[c]{@{}c@{}}Simulated\\ Modes\end{tabular}} &
  \begin{tabular}[c]{@{}c@{}}Car, bicycle, \\ pedestrian,\\ bus, train\end{tabular} &
  Car &
  \begin{tabular}[c]{@{}c@{}}Car, bicycle, \\ pedestrian,\\ bus\end{tabular} &
  \begin{tabular}[c]{@{}c@{}}Car, bicycle, \\ pedestrian\end{tabular} &
  \begin{tabular}[c]{@{}c@{}}Car, bicycle, \\ pedestrian\end{tabular} \\ \hline
\textbf{\begin{tabular}[c]{@{}c@{}}Dynamic\\ Routing\end{tabular}} &
   &
  {\checkmark} &
  {\checkmark} &
  {\checkmark} &
  {\checkmark} \\ \hline
\textbf{Parallelism} &
  \textbf{} &
  \textbf{} &
   &
   &
   \\ \hline
\textbf{Event/Time-based} &
  \begin{tabular}[c]{@{}c@{}}Discrete\\ time-based\end{tabular} &
  \begin{tabular}[c]{@{}c@{}}Discrete\\ time-based\end{tabular} &
  Event-based &
  \begin{tabular}[c]{@{}c@{}}Discrete\\ time-based\end{tabular} &
  \begin{tabular}[c]{@{}c@{}}Discrete\\ time-based\end{tabular} \\ \hline
\textbf{Agent based Model} &
   &
   &
   &
   &
  {\checkmark} \\ \hline
\textbf{\begin{tabular}[c]{@{}c@{}}Open\\ Source\end{tabular}} &
   &
   &
   &
  {\checkmark} &
  {\checkmark} \\ \hline
\textbf{Granularity} &
  Micro &
  \begin{tabular}[c]{@{}c@{}}Micro\\ /Meso\end{tabular} &
  Micro &
  \begin{tabular}[c]{@{}c@{}}Micro/\\ Meso/Macro\end{tabular} &
  \begin{tabular}[c]{@{}c@{}}Micro\\ /Meso\end{tabular} \\ \hline
\textbf{\begin{tabular}[c]{@{}c@{}}Geographic\\ Scale\end{tabular}} &
  \begin{tabular}[c]{@{}c@{}}Intersection, \\ Neighbourhood, \\ City\end{tabular} &
  \begin{tabular}[c]{@{}c@{}}City,\\ County\end{tabular} &
  \begin{tabular}[c]{@{}c@{}}City,\\ County,\\ Region\end{tabular} &
  \begin{tabular}[c]{@{}c@{}}Intersection, \\ Neighbourhood, \\ City, County\end{tabular} &
  \begin{tabular}[c]{@{}c@{}}City,\\ County\end{tabular}
   \\ \hline
\textbf{\begin{tabular}[c]{@{}c@{}}Traffic\\ Structure\end{tabular}} &
  Geographical &
  Geographical &
  \begin{tabular}[c]{@{}c@{}}Geographical/\\ Logical\end{tabular} &
  \begin{tabular}[c]{@{}c@{}}Geographical/\\ Logical\end{tabular} &
  Logical \\ \hline
\textbf{Compatibility} &
  SUMO, Aimsun &
   &
   &
  VISSIM, Aimsun &
  BEAM \\ \hline
\textbf{\begin{tabular}[c]{@{}c@{}}Parameter\\ Uncertainty\end{tabular}} &
   &
   &
   &
  vehicle and routing &
  agent behavior \\ \hline
\textbf{\begin{tabular}[c]{@{}c@{}}Simulating\\ Contingency\end{tabular}} &
   &
   &
   &
  {\checkmark} &
   \\ \hline
\textbf{\begin{tabular}[c]{@{}c@{}}Additional\\ Feature\end{tabular}} &
   &
   &
   &
  \begin{tabular}[c]{@{}c@{}}Deployment \\ on cloud\end{tabular} &
  \begin{tabular}[c]{@{}c@{}}Deployment \\ on cloud\end{tabular} \\ \hline
\end{tabular}
}
\end{table}

Apart from the points mentioned above, we also take granularity, compatibility, dynamic routing, traffic structure, parameter uncertainty, and simulating contingency into consideration. These are factors for more complicated simulations as well as robustness, but they can also cause the problem of slow computing time. In our classification, microscopic simulators simulates each vehicle individually, including its behavior, interactions with other vehicles, and reactions to the traffic environment; mesoscopic transportation simulators, on the other hand, model groups of vehicles and their flow through the network rather than individual vehicles. Compatibility means its possible interactions with other simulators. Dynamic routing refers to the methodology where the routes of vehicles are continuously updated based on real-time or near-real-time traffic conditions. For traffic structure, a "logical" structure pertains to the way data flows are organized based on rules and protocols within the network, often independent of physical locations. Conversely, a simulator with "geographical" structure means it contains the physical layout of the network, where data paths are influenced by actual geographic distances and locations of the network components. For simulating contingency, it means if the simulator has the function to respond to unexpected events or conditions. Additionally, some simulators can be deployed on cloud, which can be helpful for cooperation combined with open source. For all the simulators, we categorized them into 13 characteristics as shown in Table \ref{A Comparison Between Different Ground Simulators (Part 1)}, \ref{A Comparison Between Different Ground Simulators (Part 2)}, \ref{A Comparison Between Different Air Simulators (Part 1)} and \ref{A Comparison Between Different Air Simulators (Part 2)}.

\begin{table}[]
\caption{A Comparison Between Different Ground Simulators (Part II)}
\label{A Comparison Between Different Ground Simulators (Part 2)}
\centering
\scalebox{0.8}{
\begin{tabular}{cccccc}
\hline
\textbf{Simulator} &
  \textbf{6. BEAM \cite{sheppard2017modeling}} &
  \textbf{7. POLARIS \cite{auld2016polaris} } &
  \textbf{8. Aimsun \cite{casas2010traffic}} &
  \textbf{9. Mobiliti \cite{chan2018mobiliti}} &
  \textbf{10. MANTA \cite{yedavalli2022microsimulation}} \\ \hline
\textbf{\begin{tabular}[c]{@{}c@{}}Simulated\\ Modes\end{tabular}} &
  \begin{tabular}[c]{@{}c@{}}Car, bicycle, \\ pedestrian\end{tabular} &
  Car &
  \begin{tabular}[c]{@{}c@{}}Car, bicycle, \\ pedestrian,\\ bus, train\end{tabular} &
  Car &
  Car \\ \hline
\textbf{\begin{tabular}[c]{@{}c@{}}Dynamic\\ Routing\end{tabular}} &
  {\checkmark} &
  {\checkmark} &
  {\checkmark} &
  {\checkmark} &
   \\ \hline
\textbf{Parallelism} &
  \begin{tabular}[c]{@{}c@{}}CPU\\ Parallel\end{tabular} &
  \begin{tabular}[c]{@{}c@{}}CPU\\ Parallel\end{tabular} &
   &
  \begin{tabular}[c]{@{}c@{}}CPU\\ Parallel\end{tabular} &
  \begin{tabular}[c]{@{}c@{}}GPU\\ Parallel\end{tabular} \\ \hline
\textbf{Event/Time-based} &
  \begin{tabular}[c]{@{}c@{}}Discrete\\ time-based\end{tabular} &
  Event-based &
  \begin{tabular}[c]{@{}c@{}}Discrete\\ time-based\end{tabular} &
  Event-based &
  Event-based \\ \hline
\textbf{Agent based Model} &
  {\checkmark} &
  {\checkmark} &
   &
  {\checkmark} &
  {\checkmark} \\ \hline
\textbf{\begin{tabular}[c]{@{}c@{}}Open\\ Source\end{tabular}} &
  {\checkmark} &
  {\checkmark} &
   &
   &
  {\checkmark} \\ \hline
\textbf{Granularity} &
  \begin{tabular}[c]{@{}c@{}}Micro\\ /Meso\end{tabular} &
  Meso &
  \begin{tabular}[c]{@{}c@{}}Micro\\ /Meso\end{tabular} &
  \begin{tabular}[c]{@{}c@{}}Micro\\ /Meso\end{tabular} &
  Micro \\ \hline
\textbf{\begin{tabular}[c]{@{}c@{}}Geographic\\ Scale\end{tabular}} &
  \begin{tabular}[c]{@{}c@{}}City,\\ County\end{tabular} &
  \begin{tabular}[c]{@{}c@{}}City,\\ County\end{tabular} &
  \begin{tabular}[c]{@{}c@{}}Intersection, \\ Neighbourhood, \\ City, County\end{tabular} &
  \begin{tabular}[c]{@{}c@{}}City,\\ County\end{tabular} &
  \begin{tabular}[c]{@{}c@{}}City,\\ County\end{tabular} \\ \hline
\textbf{\begin{tabular}[c]{@{}c@{}}Traffic\\ Structure\end{tabular}} &
  Logical &
  Logical &
  \begin{tabular}[c]{@{}c@{}}Geographical/\\ Logical\end{tabular} &
  Logical &
  Logical \\ \hline
\textbf{Compatibility} &
  MATSim &
   &
  VISSIM, SUMO &
   &
  VertiSim, $Fe^3$ \\ \hline
\textbf{\begin{tabular}[c]{@{}c@{}}Parameter\\ Uncertainty\end{tabular}} &
  agent behavior &
   &
  vehicle, public transport, and incident &
   &
   \\ \hline
\textbf{\begin{tabular}[c]{@{}c@{}}Simulating\\ Contingency\end{tabular}} &
   &
  {\checkmark} &
  {\checkmark} &
   &
   \\ \hline
\textbf{\begin{tabular}[c]{@{}c@{}}Additional\\ Feature\end{tabular}} &
  \begin{tabular}[c]{@{}c@{}}Deployment \\ on cloud\end{tabular} &
   &
  \begin{tabular}[c]{@{}c@{}}Deployment \\ on cloud\end{tabular} &
  \begin{tabular}[c]{@{}c@{}}Super\\ Computer\end{tabular} &
  \begin{tabular}[c]{@{}c@{}}Deployment \\ on cloud\end{tabular} \\ \hline
\end{tabular}
}
\end{table}

\begin{table}[]
\caption{A Comparison Between Different Air Simulators (Part I)}
\label{A Comparison Between Different Air Simulators (Part 1)}
\centering
\scalebox{0.8}{
\begin{tabular}{ccccc}
\hline
\textbf{Simulator} &
  \textbf{1. Fe$^{\textbf{3}}$   \cite{xue2018Fe3}} &
  \textbf{2. AirTraf 2.0 \cite{yamashita2020newly}} &
  \textbf{3. VertiSim \cite{yedavalli2021simuam}} &
  \textbf{4. DTALite \cite{zhou2014dtalite}} \\ \hline
\textbf{\begin{tabular}[c]{@{}c@{}}Simulated\\ Modes\end{tabular}} &
  Commercial Aircraft &
  Commercial Aircraft &
  UAM &
  Commercial Aircraft \\ \hline
\textbf{\begin{tabular}[c]{@{}c@{}}Dynamic\\ Routing\end{tabular}} &
   &
   &
   &
  {\checkmark} \\ \hline
\textbf{Parallelism} &
  \begin{tabular}[c]{@{}c@{}}GPU\\ Parallel\end{tabular} &
  \begin{tabular}[c]{@{}c@{}}CPU\\ Parallel\end{tabular} &
  \begin{tabular}[c]{@{}c@{}}CPU\\ Parallel\end{tabular} &
  \begin{tabular}[c]{@{}c@{}}CPU\\ Parallel\end{tabular} \\ \hline
\textbf{Event/Time-based} &
  \begin{tabular}[c]{@{}c@{}}Discrete\\ time-based\end{tabular} &
  Event-based &
  \begin{tabular}[c]{@{}c@{}}Discrete\\ time-based\end{tabular} &
  \begin{tabular}[c]{@{}c@{}}Discrete\\ time-based\end{tabular} \\ \hline
\textbf{Agent based Model} &
   &
   &
  {\checkmark} &
  {\checkmark} \\ \hline
\textbf{\begin{tabular}[c]{@{}c@{}}Open\\ Source\end{tabular}} &
   &
   &
  \multicolumn{1}{l}{} &
  {\checkmark} \\ \hline
\textbf{Granularity} &
  Micro &
  \begin{tabular}[c]{@{}c@{}}Micro\\ /Meso\end{tabular} &
  Micro &
  Meso \\ \hline
\textbf{\begin{tabular}[c]{@{}c@{}}Geographic\\ Scale\end{tabular}} &
  \begin{tabular}[c]{@{}c@{}}Airport,\\ Region\end{tabular} &
  \begin{tabular}[c]{@{}c@{}}City,\\ County\end{tabular} &
  \begin{tabular}[c]{@{}c@{}}Airport,\\ Region\end{tabular} &
  \begin{tabular}[c]{@{}c@{}}Airport,\\ Region\end{tabular} \\ \hline
\textbf{\begin{tabular}[c]{@{}c@{}}Traffic\\ Structure\end{tabular}} &
  Logical &
  Logical &
  Logical &
  Logical \\ \hline
\textbf{Compatibility} &
  MANTA, VertiSim &
   &
  MANTA, Fe3 &
   \\ \hline
\textbf{\begin{tabular}[c]{@{}c@{}}Parameter\\ Uncertainty\end{tabular}} &
  Wind &
   &
  \multicolumn{1}{l}{} &
  vehicle depart and arrive \\ \hline
\textbf{\begin{tabular}[c]{@{}c@{}}Simulating\\ Contingency\end{tabular}} &
   &
   &
  \multicolumn{1}{l}{} &
  {\checkmark} \\ \hline
\end{tabular}
}
\end{table}

\begin{table}[]
\caption{A Comparison Between Different Air Simulators (Part II)}
\label{A Comparison Between Different Air Simulators (Part 2)}
\centering
\scalebox{0.8}{
\begin{tabular}{cccccc}
\hline
\textbf{Simulator} &
  \textbf{5. FACET \cite{bilimoria2001facet}} &
  \textbf{6. TESTBED \cite{robinson2017development}} &
  \textbf{7. Bluesky \cite{hoekstra2016bluesky}} &
  \textbf{8. AEDT \cite{roof2007aviation}} &
  \textbf{9. AEIC} \cite{simone2013rapid}\\ \hline
\textbf{\begin{tabular}[c]{@{}c@{}}Simulated\\ Modes\end{tabular}} &
  Commercial Aircraft &
  Commercial Aircraft &
  \begin{tabular}[c]{@{}c@{}}Commercial Aircraft,\\ UAM\end{tabular} &
  Commercial Aircraft &
  Commercial Aircraft \\ \hline
\textbf{\begin{tabular}[c]{@{}c@{}}Dynamic\\ Routing\end{tabular}} &
  {\checkmark} &
  {\checkmark} &
  {\checkmark} &
   &
   \\ \hline
\textbf{Parallelism} &
   &
   &
   &
   &
  CPU Parallel \\ \hline
\textbf{Event/Time-based} &
  Event-based &
  Event-based &
  \begin{tabular}[c]{@{}c@{}}Discrete\\ time-based\end{tabular} &
  \begin{tabular}[c]{@{}c@{}}Discrete\\ time-based\end{tabular} &
  Event-based \\ \hline
\textbf{Agent based Model} &
   &
   &
   &
   &
   \\ \hline
\textbf{\begin{tabular}[c]{@{}c@{}}Open\\ Source\end{tabular}} &
  {\checkmark} &
   &
  {\checkmark} &
   &
   \\ \hline
\textbf{Granularity} &
  Meso &
  Meso &
  \begin{tabular}[c]{@{}c@{}}Micro\\ /Meso\end{tabular} &
  Macro &
  Marco \\ \hline
\textbf{\begin{tabular}[c]{@{}c@{}}Geographic\\ Scale\end{tabular}} &
  \begin{tabular}[c]{@{}c@{}}City,\\ County, \\ National Scale\end{tabular} &
  \begin{tabular}[c]{@{}c@{}}City,\\ County\end{tabular} &
  \begin{tabular}[c]{@{}c@{}}City,\\ County\end{tabular} &
  \begin{tabular}[c]{@{}c@{}}City,\\ County, \\ Country\end{tabular} &
  \begin{tabular}[c]{@{}c@{}}County, \\ Country\end{tabular} \\ \hline
\textbf{\begin{tabular}[c]{@{}c@{}}Traffic\\ Structure\end{tabular}} &
  Logical &
  Logical &
  Logical &
  Logical &
  Logical \\ \hline
\textbf{\begin{tabular}[c]{@{}c@{}}Parameter\\ Uncertainty\end{tabular}} &
   &
   &
   &
   &
  \begin{tabular}[c]{@{}c@{}}Aircraft Altitude, \\ Weight\end{tabular} \\ \hline
\textbf{\begin{tabular}[c]{@{}c@{}}Simulating\\ Contingency\end{tabular}} &
   &
   &
  {\checkmark} &
   &
   \\ \hline
\end{tabular}
}
\end{table}

\subsubsection{Influential Factors in the Evolution of Simulators and Their Direct Impact on UAM Integration}
\begin{figure}[!htbp]

\centerline{\includegraphics[width=1\linewidth]{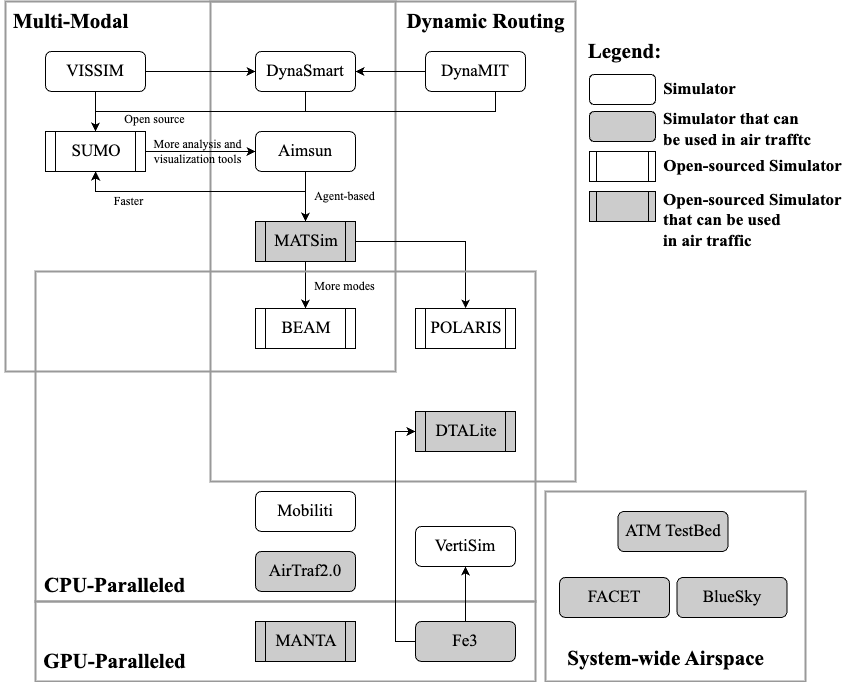}}
\caption{Simulators Evolution Diagram}
\label{Simulators Evolution Diagram}
\end{figure}

The 17 traffic simulators are evaluated in this paper, each exhibiting unique characteristics and capabilities in the field of transportation modeling. Figure \ref{Simulators Evolution Diagram} depicts these 17 simulators, organized by their defining features.

Primarily, these simulators are classified based on their capacity to simulate air or ground traffic. Subsequently, their capabilities are further examined in relation to two key functionalities of any transportation simulator - multi-modal operation and dynamic routing. Historically, discrete-event simulation and network flow simulation were the standard techniques used. However, with the advent of multi-modal simulators and the concurrent challenges they posed, agent-based simulation was introduced \cite{w2016multi}. The integration of these functions tends to complicate the model, often resulting in slower processing speeds.

One solution to manage large data sets effectively is the use of supercomputers as Mobiliti did \cite{chan2018mobiliti, 9925762}.  On the other hand, as urban scales expanded, there was a pivot towards harnessing the power of parallel computation, first with CPU and later extending to the more robust capabilities of GPU parallel computation as MANTA did \cite{yedavalli2021microsimulation}. 

The use of GPU parallel computation is necessary due to the sheer volume of data involved in simulating at a regional scale. GPUs are particularly adept at handling computations that are repetitive but not heavily logic-intensive. This makes them ideal for microscopic simulations and those that are time-driven. With the introduction of UAM, the granularity required for simulations becomes even more critical. To ensure the highest levels of accuracy and safety, it is imperative to lean towards microscopic simulations. Furthermore, UAM offers a high degree of freedom in its flight paths, making it essential to use more detailed simulations to discern the interplay between aerial and ground traffic. This level of detail is paramount in understanding the intricate dynamics and potential bottlenecks that might arise when integrating UAM into the existing urban fabric.

\section{Key Research Gaps and Future Directions}

UAM represents a transformative paradigm in urban transportation, promising to alleviate congestion, reduce environmental impact, and enhance mobility within metropolitan areas. While significant strides have been made in UAM research, there remain critical research gaps and promising future directions that warrant comprehensive exploration and attention.

\subsection{Comprehensive Regional-Scale Simulation}

In the field of UAM, the development of advanced regional-scale simulation tools is a critical research priority. Current simulators are insufficient for the complex dynamics of UAM in urban settings. Future efforts should focus on creating a simulator specifically for UAM integration, capable of handling its multifaceted aspects in multi-modal transportation systems. Additionally, research must address the environmental impact of UAM, especially in terms of greenhouse gas emissions. This requires more precise data and advanced simulation models for accurate environmental assessments. The proposed simulator would be pivotal in enabling detailed evaluations of UAM's impact on urban transportation and aiding policymakers in optimizing UAM integration\cite{johnson2022nasa}\cite{yun2020requirement}.

\subsection{Holistic Impact Evaluation}

Understanding the holistic impact of UAM on large-scale urban traffic and infrastructure is paramount for achieving the full potential of this innovative mode of transportation. Future research endeavors should center around simulation-based impact evaluation, which encompasses a comprehensive assessment of UAM's effects on various critical factors:

\begin{itemize}
    \item Infrastructure Needs: Evaluation of necessary infrastructure enhancements and the associated costs to support a growing UAM network. This includes vertiport locations, maintenance facilities, and airspace management systems.
    \item Socio-economic Impacts: Investigation into the broader economic benefits or challenges UAM might introduce, from job creation to potential shifts in property values due to vertiport proximities.
    \item Integration with Existing Transit: Analyzing seamless intermodal transfers, ensuring that UAM complements and efficiently links with other transportation modes, like subways, buses, or rideshares.
    \item Public Perception and Acceptance: Understanding community sentiments about UAM, from concerns about privacy to general enthusiasm or resistance. This will play a crucial role in the adoption and smooth integration of UAM.
    \item Regulatory and Policy Implications: Delving into the potential changes in urban transport policies, aviation regulations, and how these might evolve to accommodate UAM.
    \item Urban Aesthetics and Cityscape: Assessing the visual impact of UAM on the city's skyline, including the design and placement of vertiports and how UAM operations might change the aesthetics of urban spaces.
\end{itemize}

Efforts should be directed towards establishing a unified framework for conducting these assessments, allowing for cross-city comparisons and ensuring that UAM deployment aligns with the unique characteristics and challenges of each urban environment.

\subsection{Collaborative Research and Development}

While UAM holds immense promise for urban transportation, it also underscores the necessity for concerted research and development efforts. Researchers, policymakers, industry stakeholders, and urban planners must collaborate to overcome the multifaceted challenges associated with UAM integration.

Research consortia should focus on fostering innovation in technology, policy, and infrastructure. This includes:

\begin{itemize}
    \item Technological Advancements: Investing in cutting-edge UAM technologies, including vehicle design, airspace management, and autonomous systems, to enhance efficiency, safety, and sustainability.
    \item Policy Frameworks: Developing forward-thinking regulatory and policy frameworks that strike a balance between fostering innovation and safeguarding public interests, including safety, privacy, and equity.
    \item Infrastructure Adaptation: Adapting urban infrastructure to accommodate UAM, including the design of vertiports, charging infrastructure, and seamless intermodal connections.
    \item Stakeholder Engagement: Encouraging active involvement of communities and stakeholders to ensure that UAM implementation aligns with local needs and aspirations.
\end{itemize}
\section{Challenges and conclusions}

\label{Challenges and conclusions}

In conclusion, this research provides an in-depth survey of the current state of UAM integration into urban transport systems through advanced simulation techniques. Our analysis indicates that while UAM has the potential to significantly enhance mobility in metropolitan areas, existing traffic simulators are inadequate for comprehensively representing UAM scenarios. These simulators either focus too narrowly on vehicle performance and airspace management or are overly simplistic, failing to capture the complex dynamics of UAM.

The study identifies key areas for further research and development, emphasizing the necessity of an innovative simulator tailored for UAM. Such a simulator would enable a detailed assessment of UAM’s impact on urban traffic patterns, travel time, noise, emissions, and safety. It would also address the broader implications of UAM on existing on-demand ground transportation services and raise important considerations about equity and ground-side safety.

We conclude that the integration of UAM into urban transportation is promising but requires a targeted approach in simulation technology to fully harness its benefits. The future of UAM relies on the development of comprehensive regional simulators that can accurately model its influence on city transportation networks, ensuring informed decision-making that considers not just technical performance but also societal and environmental impacts.

\section{Statements}
This research is supported in part by Supernal contract number 052838-002.

\newpage

\bibliography{sample}

\end{document}